\documentclass[11pt, dvipsnames]{wlscirep}
\usepackage[utf8]{inputenc}
\usepackage[T1]{fontenc}
\usepackage{lineno}
\usepackage{cleveref}
\usepackage{pdfpages}
\modulolinenumbers[2]
\usepackage[nolist,nohyperlinks]{acronym}
\acrodef{XAI}{explainable artificial intelligence}
\acrodef{LIME}{local interpretable model-agnostic explanations}
\acrodef{SHAP}{Shapley additive explanations}
\acrodef{CO}{causal ordering}
\acrodef{DAG}{directed acyclic graph}
\acrodef{BDT}{boosted decisions tree}
\acrodef{ACS}{American Community Survey}

\title{Socioeconomic disparities and COVID-19: the causal connections}
\author[a,*]{Tannista Banerjee}
\author[b,c,*]{Ayan Paul}
\author[d,e,*]{Vishak Srikanth} 
\author[f,g,*]{Inga Str\"umke}

\affil[a]{Department of Economics, Auburn University, 140 Miller Hall, Auburn, AL, 36849, USA}
\affil[b]{DESY, Notkestra{\ss}e 85, D-22607 Hamburg, Germany}
\affil[c]{Institut f\"ur Physik, Humboldt-Universit\"at zu Berlin, D-12489 Berlin, Germany}
\affil[d]{BASIS Independent Silicon Valley, San Jose, CA}
\affil[e]{Stanford Online High School, Stanford, CA}
\affil[f]{Department of Engineering Cybernetics, NTNU, 7034 Trondheim, Norway}
\affil[g]{Department of Holistic Systems, SimulaMet, 0167 Oslo, Norway}

\affil[*]{e-mail: 
\href{mailto:tzb0018@auburn.edu}{tzb0018@auburn.edu}
\href{mailto:ayan.paul@desy.de}{ayan.paul@desy.de}, 
\href{mailto:ivishak@gmail.com}{ivishak@gmail.com}, 
\href{mailto:inga@simula.no}{inga.strumke@ntnu.no}, 
}

\begin{abstract}
The analysis of causation is a challenging task that can be approached in various ways. With the increasing use of machine learning based models in computational socioeconomics, explaining these models while taking causal connections into account is a necessity. In this work, we advocate the use of an explanatory framework from cooperative game theory augmented with $do$ calculus, namely causal Shapley values. Using causal Shapley values, we analyze socioeconomic disparities that have a causal link to the spread of COVID-19 in the USA. We study several phases of the disease spread to show how the causal connections change over time. We perform a causal analysis using random effects models and discuss the correspondence between the two methods to verify our results. We show the distinct advantages a non-linear machine learning models have over linear models when performing a multivariate analysis, especially since the machine learning models can map out non-linear correlations in the data. In addition, the causal Shapley values allow for including the causal structure in the variable importance computed for the machine learning model.
\end{abstract}
\begin{document}
% \linenumbers
\flushbottom
\maketitle

\thispagestyle{fancy}
\rhead{DESY 21-150 $\left.\middle|\right.$ HU-EP-21/33}

%#####################
\section*{Introduction}
%#####################

The early stages of the spread of COVID-19 in the USA laid bare how socioeconomic disparities bring about a disproportionate spread of the disease in certain parts of society~\cite{MILLETT2020,10.1001/jama.2020.6548,doi:10.1056/NEJMp2021971,10.15585/mmwr.mm6933e1,10.1001/jama.2020.11374,DIMAGGIO20207,doi:10.1080/13557858.2020.1853067,Pareek2020,Laurencin2020,Goyale2020009951,UchicagoWP,Weill19658,Qiu2020,2020arXiv200407947S}. This trend is a pattern that has been historically observed for several diseases caused by viruses like HIV~\cite{Ransome2016}, MERS-CoV, SARS-CoV, Ebola~\cite{Farmer1996,10.1371/journal.pone.0012763,Quinn2014} etc. Even beyond variances in socioeconomic conditions, ethnicity is a contributing factor~\cite{NAP12875,10.1001/jama.1990.03440170066038,PMID:11110355,PMID:9386949} for disparities in healthcare. A good understanding of the causal connection between socioeconomic disparities and the spread of a disease is necessary for implementing policies to mitigate disease spread amongst those who are the most vulnerable.

In a recent work, it was shown that the spread of COVID-19 is correlated to various socioeconomic metrics averaged at the county level~\cite{10.1088/2632-072X/ac0fc7} during the initial stages of the pandemic. The use of Shapley values was introduced with a machine learning framework to understand how socioeconomic disparities affect the spread of the disease. However, Shapley values decomposing a machine learning model probe the correlations between the endogenous and exogenous variables and do not the the causal connection in the data into account. In this work, we go a step further and introduce causal Shapley values to the analysis of disease spread. 

We obtain a regression model by training a machine learning model in a supervised manner on COVID-19 prevalence data. As the resulting model is too complex to be considered interpretable, we use methods from the field of explainable AI to explain the model's predictions. Specifically, we use a framework built on the game-theoretic solution concept of Shapley values~\cite{Shapley1953}, upon which the packages of \ac{SHAP}~\cite{Lundberg2017} and shapr~\cite{Loland2019} are based. The idea behind these was recently built upon by Ref.~\cite{Heskes2020} to avoid independence assumptions in the data as well as to incorporate causal knowledge from causal chain graphs in the Shapley value calculation. Our objectives in this work can be delineated as:
\begin{itemize}
\itemsep0em
\item We study the spread of COVID-19 in three different regions of the USA broken down by the pattern of spread of the disease.
\item We build socioeconomic metrics using data sources from the US Census Bureau at the county level to model the demographic distribution of the population.
\item We use an ensemble of boosted decisions trees (BDTs) to build a regression model of the data.
\item We calculate the causal Shapley values using these regression models to understand the causal connections between the socioeconomic metrics and the spread of COVID-19.
\item We study two different times, the early stage the disease spread between February and July 2020 and the later stage of the disease spread between July 2020 and January 2021 to understand the variance in the causal connections.
\item We compare the multivariate methods that we establish here with linear random effects models and discuss the relative advantage of using machine learning for highly correlated datasets.
\end{itemize}

To avoid the effects of causation being washed out due to large scale averaging~\cite{10.1088/2632-072X/ac0fc7}, we divide our analysis into three regions. This is particularly important since the demographic nature of various parts of the USA are different. In addition, the pattern of disease spread were also observed to vary in different parts of the USA. The states with higher densities were significantly affected in the first wave of the pandemic while the southern states were significantly affected by the second wave. The states on the west coast were affected throughout the extent of the pandemic. Demographically, the northeastern states have several urban areas that are closely clustered together. However, in the southern states the urban areas are spread far apart and a large fraction of the population lives in rural areas where the socioeconomic conditions are different from the urban areas. The states on the west coast are a mix of urban and rural areas with a very different economic structure. Hence, we will focus on three different clusters of states.

\begin{itemize}
\itemsep0em
    \item {\bf High population density regions:} States with population density over 400 individuals per sq. km: District of Columbia, New Jersey, Rhode Island, Massachusetts, Connecticut, Maryland, Delaware and New York.
    \item {\bf The southern states:} These include: Alabama, Arkansas, Florida, Georgia, Kentucky, Louisiana, Mississippi, North Carolina, Oklahoma, South Carolina, Tennessee, Texas, Virginia and West Virginia.
    \item {\bf The west coast:} These include: California, Oregon and Washington.
\end{itemize}

We hope that by highlighting how the socioeconomic disparities that aggravated the spread of COVID-19 in of the community, and often amongst those who are at a social disadvantage, we can not only aid in future policy building but also establish a robust method for causal analysis that can be easily used for highly multivariate datasets. 

% --------------------
\section*{Causal ordering}
% --------------------
The framework advocated by Heskes et al.~\cite{Heskes2020} for causal Shapley values accounts for indirect effects in order to take the causal structure of the data into account in when estimating the Shapley values. The approach recognizes that it is not practical to compute interventional probabilities to account for indirect effects and instead proposes a causal chain graph The (partial) causal ordering in the data is represented as a causal chain graph by a \ac{DAG}. A \ac{CO} is then a nested list that maps to a sequence of the respective causal orderings of the socioeconomic metrics. Variables of equivalent causal importance are grouped into the same sub-list within the list. For example, [NW, [SC, Emp]] would represent non-white as the most significant cause which influences the next set of variables, here Senior Citizen and Employed, each with equal importance\footnote{The abbreviations and definitions of the metrics can be found in the Materials and Methods section}. More intuitively, the causal ordering depends on a hypothesis of how the socioeconomic metrics form subgroups that are causally connected. Within a subgroup, the metrics are assumed to have no causal connection but can be cyclically connected or share confounding variables. There is no fixed prescription for constructing these partial causal orderings. Rather, domain knowledge has to be used to assume how each metric can affect the others to build causal orderings. We investigate six different causal orderings, \ac{CO}\#1 - CO\#3 described below and \ac{CO}\#4 - CO\#6 included in the Supplementary Materials, and explain the rationale behind each choice. 

\begin{figure}
    \centering
    \includegraphics[width=0.7\textwidth]{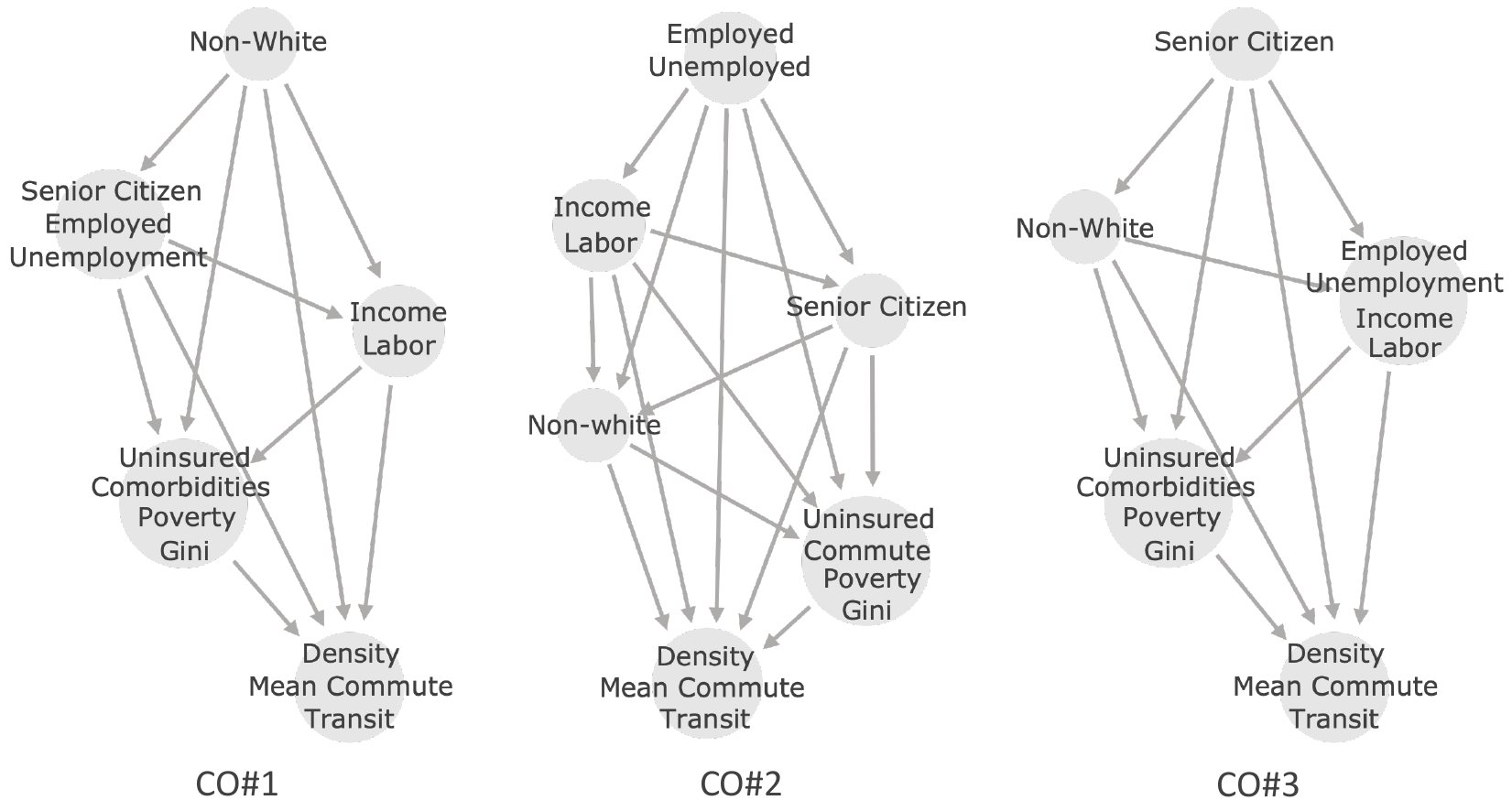}
    \caption{The three primary partial causal ordering that we consider in this work.}
    \label{fig:causal_ordering}
\end{figure}

\paragraph{CO\#1:} The first causal ordering we construct is [NW, [SC, Emp, Uemp], [Inc, Lab], [Uins, Com, Pov, GI],[Den, MC, Tran]]. The rationale for this baseline ordering is as follows.
\begin{itemize}
\itemsep0em
\item Areas with a higher proportion of non-white population tend to have higher unemployment rates and a lower proportion of employed workers~\cite{Immergluck2018}.
\item Areas with higher unemployment rates may have higher proportions of senior citizens if there is low economic mobility and younger generations move out of the area~\cite{Mills2001}.
\item Areas with higher unemployment and lower employment tend to have more people working in construction, service, delivery or production or are rust belt areas that have a declining manufacturing industry~\cite{Charles2019,Lambert2017}.
\item Areas with higher unemployment tend to be poorer and have lower incomes.~\cite{DeFina2004} 
\item Areas with more senior citizens who are on medicare or those with higher incomes individuals who receive insurance from their employer tend to have lesser uninsured fraction of the population~\cite{Finkelstein2008,Cunningham2001}.
\item Counties with higher average incomes tend to have more income inequality and a higher Gini index~\cite{Levernier1998,Nielsen1997}.
\item Areas with a higher percent of people working in construction, service, delivery, or production may have more poverty~\cite{Adelman1999}.
\item Counties with more poverty are likely to have less single family homes and more areas zoned for multi-family homes or are likely to contain inner cities and thus have a greater density~\cite{Rothwell2010,Kasarda1993}. In such dense areas, there is likely to be shorter commutes and better transit due to these areas being more populous~\cite{Bertaud04transitand,LEVINSON1997}.
\end{itemize}
% 2 WAS 4
\paragraph{CO\#2:} The second causal ordering is [[Emp, Uemp], [Inc, Lab], SC, NW, [Uins, Com, Pov, GI], [Den, MC, Tran]]. This causal ordering has unemployment, employed, income per capita, and fraction working in manufacturing or manual labor as causing the proportion of senior citizens and non-white fraction of the population. This causal ordering takes into account that senior citizens are more likely to stay in areas with a declining manufacturing industry and little economic mobility~\cite{Mills2001}. Since senior citizens are less likely to be non-white, a higher fraction of senior citizens causes there to be a lower proportion of non-whites~\cite{Blanger2016}.
%3 WAS 5
\paragraph{CO\#3:} The third causal ordering is [SC, NW, [Emp, Uemp, Inc, Lab], [Uins, Com, Pov, GI], [Den, MC, Tran]]. This causal ordering has the proportion of senior citizens causing the proportion of non-whites. Since senior citizens are less likely to be non-white, a higher fraction of senior citizens causes there to be a lower proportion of non-whites~\cite{Blanger2016}. This causal ordering also has income per capita, the proportion of people in a county working in professions classified as Labor, unemployment, and employment on equal footing. This is the case since manual labor jobs are less temporary and pay lower wages~\cite{Ono2013}.

\begin{figure*}
    \centering
    \includegraphics[width=\textwidth]{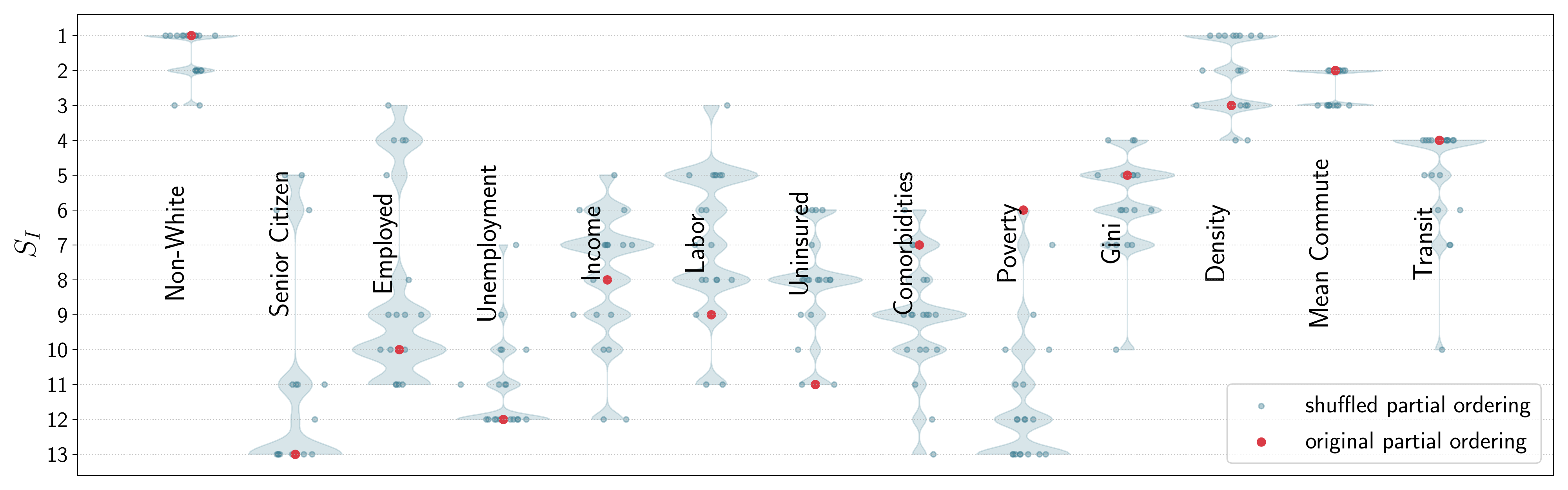}
    \caption{\it The Shapley Index, $S_I$, of all the metrics from 20 different causal ordering where the ordering of the subgroups were randomized starting from the causal ordering used in the left panel. The plot shows that the variation in $S_I$ is small for the most important metrics with high $\overline{|S_v|}$ while larger variations in $S_I$ can be seen in the less important metrics depending on the causal ordering. For more detail please see the text.}
    \label{fig:variations}
\end{figure*}

Evidently, the decision on which causal ordering to choose is quite subjective and requires some prior knowledge of how the various metrics interact with each other. In the case of the problem we are addressing, the causal ordering can even change from region to region depending on the nature of the underlying socioeconomic fabric that governs the social structure. Thus it is important to check how robust the conclusion of our work is vis-\`a-vis a perturbation in the causal ordering. Ideally, the Shapley Index, $S_I$\footnote{Shapley Index is defined in the Materials and Methods section.}, should depend not only on the data but also on the causal ordering. However, $S_I$ should not be agnostic of either the data or the causal ordering since, after all, $S_I$ should encapsulate the information from the causal ordering {\em and} the data. 

In~\autoref{fig:variations}, we show the result of shuffling the causal ordering CO\#1. We randomly rearranging the subgroups into 20 different permutations. The violin plots show the variations in $S_I$ for each metric. The red dot corresponds to the $S_I$ for CO\#1. We see that the three most important metric--non-white, density, and mean commute--do not show large variations in $S_I$. The metrics that are less causally connected to the confirmed case rate, however, show large variation. Nevertheless, there is a clear distinction between the three most important ones and the less important ones. This result shows that while $S_I$ does depend on the causal ordering, the most important metrics are tied to the information that is gleaned from the data. If we restrict ourselves to picking the metric with the highest $S_I$, and hence, the largest causal connection to the outcome, small perturbations in the causal ordering will not alter our results. This observation displays the robustness of our approach making it not as subjective as it might appear to be on the surface.

%#####################
\section*{Results}
%#####################
\begin{figure*}
    \centering
    \includegraphics{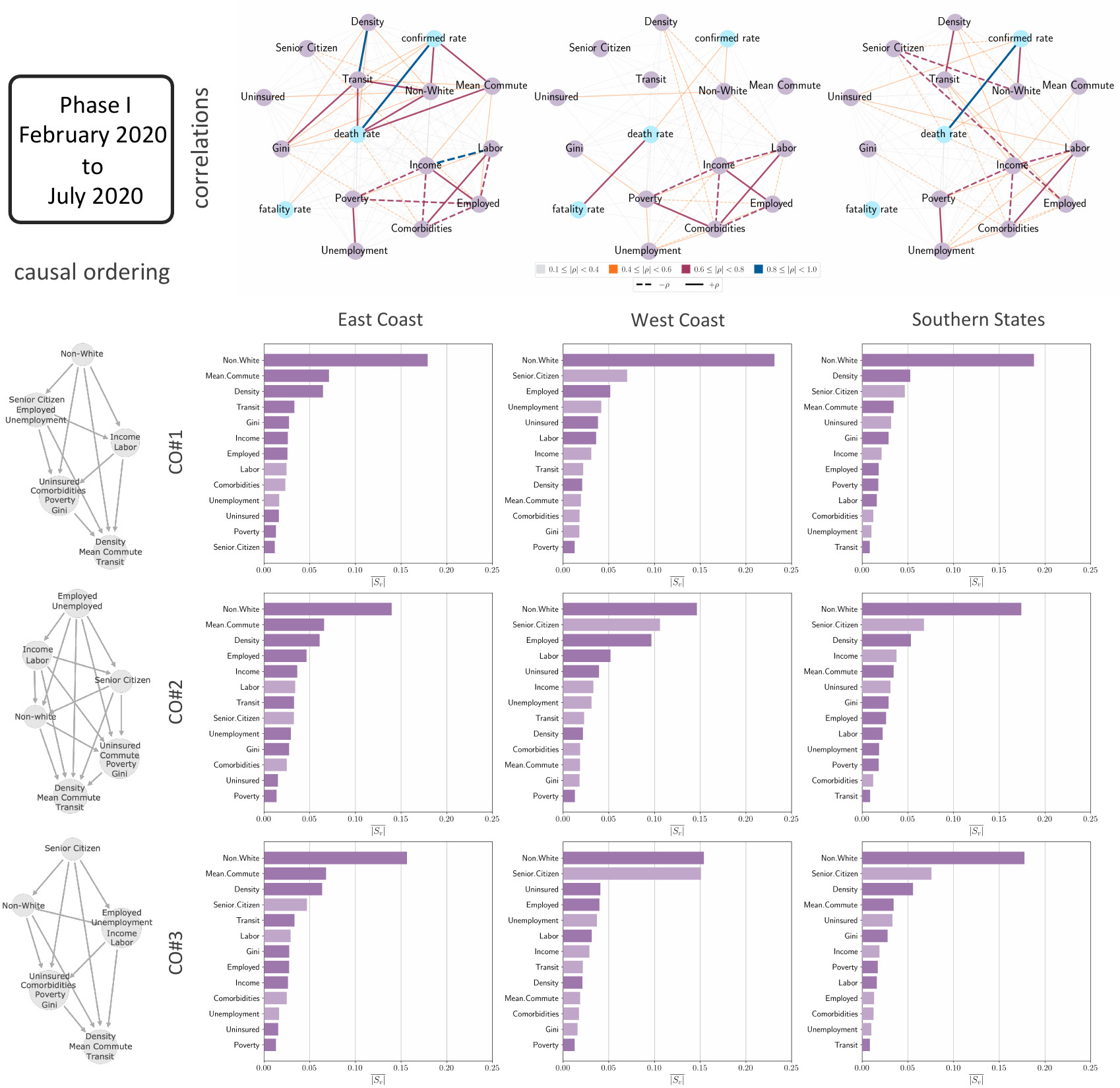}
    \caption{\it The causal connection between the socioeconomic metrics that we consider in this work and confirmed case rates in various regions of the USA from February 2020 to July 2020. The states included in the East Coast region are District of Columbia, New Jersey, Rhode Island, Massachusetts, Connecticut, Maryland, Delaware and New York. The ones in the West Coast region are California, Oregon and Washington. The states in the Southern States region are Alabama, Arkansas, Florida, Georgia, Kentucky, Louisiana, Mississippi, North Carolina, Oklahoma, South Carolina, Tennessee, Texas, Virginia and West Virginia. The networks in the top row show the correlations between the different metrics in the three regions. The graphs in the left column of the plot show the three partial causal orderings that we consider in this analysis. Each panel of the array of bar plots show the hierarchy of the metrics in terms of mean absolute causal SHAP values, $\overline{|S_v|}$, for a particular region and causal ordering. The lighter shaded bars indicate a negative effect of the metric on the confirmed case rates while the darker bars indicate a positive effect.}
    \label{fig:money-phase-I}
\end{figure*}

\begin{figure*}
    \centering
    \includegraphics{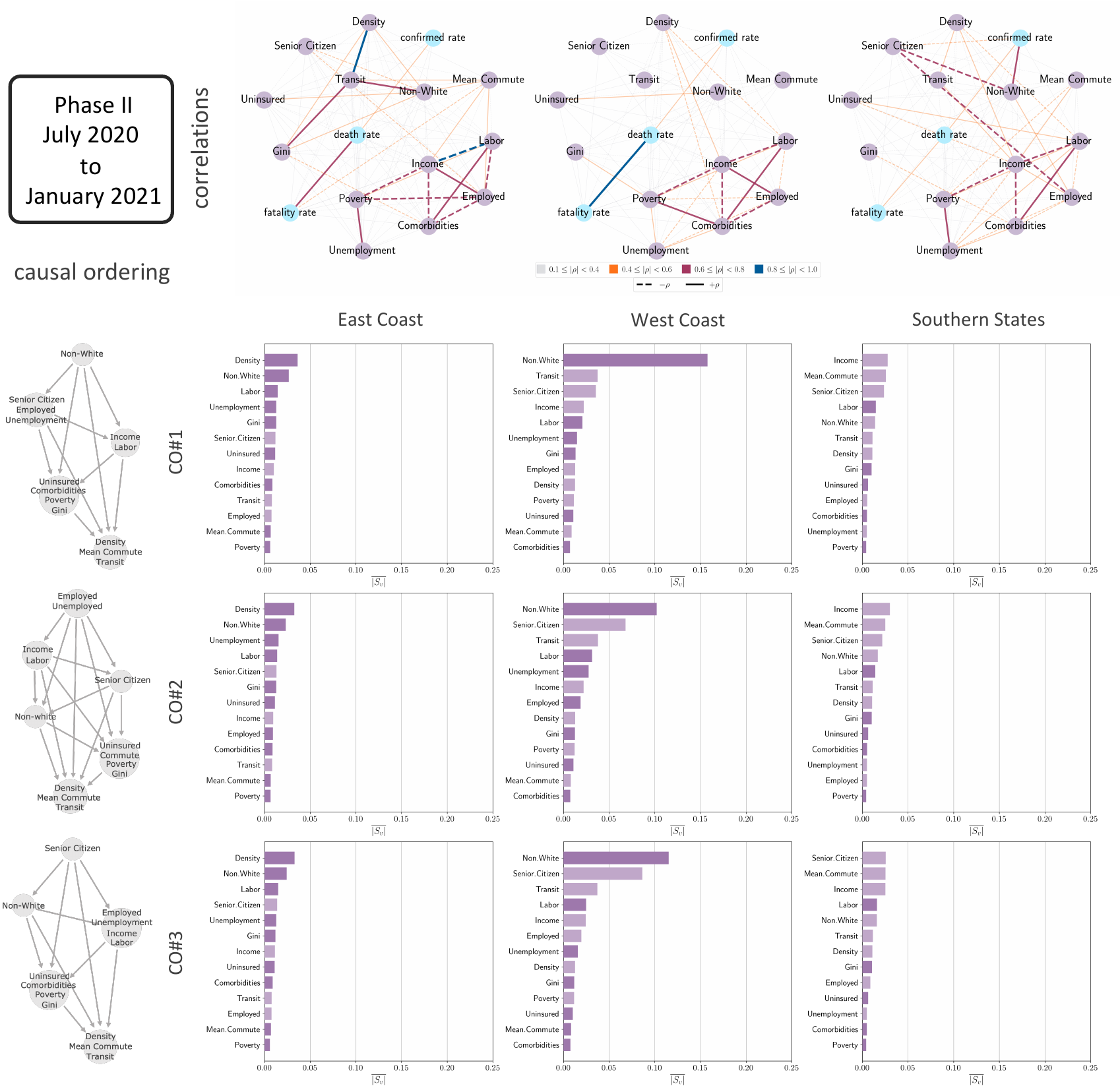}
    \caption{\it The causal connection between the socioeconomic metrics that we consider in this work and confirmed case rates in various regions of the USA for July 2020 to January 2021. The states included in the East Coast region are District of Columbia, New Jersey, Rhode Island, Massachusetts, Connecticut, Maryland, Delaware and New York. The ones in the West Coast region are California, Oregon and Washington. The states in the Southern States region are Alabama, Arkansas, Florida, Georgia, Kentucky, Louisiana, Mississippi, North Carolina, Oklahoma, South Carolina, Tennessee, Texas, Virginia and West Virginia. The networks in the top row show the correlations between the different metrics in the three regions. The graphs in the left column of the plot show the three partial causal orderings that we consider in this analysis. Each panel of the array of bar plots show the hierarchy of the metrics in terms of mean absolute causal SHAP values, $\overline{|S_v|}$, for a particular region and causal ordering. The lighter shaded bars indicate a negative effect of the metric on the confirmed case rates while the darker bars indicate a positive effect.}
    \label{fig:money-phase-II}
\end{figure*}

The question that we would like to pose is: are there any casual connection between the socioeconomic metrics and the spread of the disease and is this causal connection stable over time and geographies?

We will explore two slices of time in this section. The first one is the period when the disease had just started spreading through the population. In the beginning of 2020, COVID-19 affected the most densely populated parts of the USA, namely, the states in the east coast and some states in the west coast. However other parts of the USA, especially the rural parts of the USA, remained unaffected by large. It was the second wave in mid-2020 that affected the parts that were spared by the first wave with vast majority of the southern states seeing a surge in the number of infected individual. So a study of socioeconomic conditions that might have affected the differential advent of COVID-19 across the social strata should include the months from February 2020 to July 2020. We shall refer to this as Phase I of our study. Consequently, we shall deem Phase II as the next phase of the disease spread from July 2020 to January 2021 keeping it safely before when vaccinations could possibly have any effects. We will contrast Phase I against Phase II to understand the role socioeconomic disparities played in the spread of COVID-19.

In \autoref{fig:money-phase-I} we show the correlations and causation between the socioeconomic factors, population density and the confirmed case rate during Phase I of the disease spread. The network plots in the top panel display the linear correlations between the exogenous and endogenous variable(s). From the correlations in the plots being distinct for each region, one can expect the causal connections between the variables should be different too. This is indeed confirmed by the bar plots, the columns corresponding to the different regions and the rows corresponding to the different causal ordering. A very clear pattern emerges independent of the causal ordering. In the East Coast region during Phase I, the primary causation is driven by the fraction of non-white population, the mean commute and the population density. While the fraction of non-white population has by far the largest causal connection with the confirmed case rates, the mean commute and density variables interchange positions depending on the causal ordering used. However, it is clear that these three are the most important causation. 

In the West Coast region, the fraction of non-white and the fraction of senior citizen in a county are the top causation. The fraction employed in a county complete the top three list in the CO\#1 and CO\#2 while loses importance in CO\#3 leading us to believe that its importance is more sensitive to the causal ordering. In the Southern States the conclusion is similar with the fraction of non-white population, the fraction of senior citizen and the population density having the strongest causal connections with the confirmed case rate with non-white fraction being the primary causation.

Moving on to Phase II the pattern is quite different. In \autoref{fig:money-phase-II} we see that the value of $\overline{|S_v|}$ for all the variables are much lower than the corresponding ones in Phase I for both the East Coast States and the Southern States leading us to conclude that the variables have less effect in causing the spread of the disease which is more homogeneous geographically. The case for the West Coast region is somewhat different with the non-white fraction still holding some causal connection with the spread of the disease along with the fraction of senior citizen for CO\#2 and CO\#3. In addition, the correlation patterns shown in the network plots in \autoref{fig:money-phase-II} differ from those in \autoref{fig:money-phase-I} and seem to be more sparse for the East Coast region and the Southern States while being relatively unchanged for the West Coast region.

%#####################
\section*{Econometric analysis}
%#####################
The machine learning framework along with the causal Shapley values are novel method for analyzing causalities in socioeconomic data. To validate its strengths and to highlight its differences with a more traditional linear causal analysis we use a simple linear model to analyze the effect of socioeconomic metrics on the spread of COVID-19 across the USA. 

We use weekly balanced panel data, and we safely assume that standard errors in the balanced panel data are not independent. To begin with, we perform the White test and Breusch–Pagan test to detect the presence of heteroscedasticity. For each of the six datasets (three regions and two phases) we get a $p$-value $< 0.05$ from both the tests which leads us to reject the null hypothesis of homoscedasticity and we conclude that heteroscedasticity is present in the residual of this empirical model specification. Furthermore, we perform the Durbin-Watson test for auto-correlation and get test results of < 1 for all datasets leading us to the conclusion that there are significant positive auto-correlations in the datasets. Therefore, we need to use a random effect model with clustered standard error at the county level. These clustered standard errors allow for the presence of heteroscedasticity and correlation in the error term within a cluster. A fixed effect model cannot be used given the socioeconomic metrics do not change over the times during which the COVID-19 prevalence data is collected. The random effect model estimates the effects of time-invariant socioeconomic metrics as presented in the data collected from the US Census Bureau. We build the model with only a subset of the socioeconomic metrics that we consider in the previous analysis since highly correlated metrics cannot be used to fit a linear random effect model. Hence we chose four of the metrics that are not highly correlated and perform the analysis so as to reasonably compare it with the results obtained from the machine learning framework.

The result of the empirical analysis for Phase I is presented in \autoref{tab:ecotab1}. The dependent variable is the $\log_{10}$ of weekly COVID-19 confirmed cases per 100,000 individuals within a county. The first row of \autoref{tab:ecotab1} shows that for all the regions, higher population density increases the probability of the spread of the number of COVID-19 cases in the region. The effect is positive and significant ($p$-value $<0.05$). The second row shows that in regions with higher unemployment rate, the spread of COVID-19 is relatively less as seen in the numbers from the west coast ($p$-value $<0.05$). Third and fourth row have the expected results that higher income region and regions with a smaller fraction of non-white population sees a decrease in the probability of the spread of COVID-19 ($p$-value $<0.05$). The effect is significant for all three regions.

The Phase II results in the lower half of \autoref{tab:ecotab1} has similar results as Phase I and shows that socioeconomic metrics affect the spread of COVID-19 significantly, especially in the West Coast states. The primary difference between Phase I and Phase II is that some of the effects of the socioeconomic metrics on the spread of COVID-19 decreases in Phase II compared to Phase I as presented in our previous analysis, especially for the East coast and the Southern States regions. These results highlight two implications of the causal analyses. Firstly, the initial health care policy intervention of some of the US states and the federal government seem to have helped in reducing the differential in the spread of the pandemic from what was seen at the initial stage and was disproportionately affecting the more vulnerable. Secondly, socioeconomic metrics prove to be important in shaping the possible policy interventions to control the spread of the pandemic.

\begin{table}
    \centering
    \begin{tabular}{l|ccc}
         \toprule
         \multicolumn{4}{c}{Phase I}\\
         \hline
         Variables &  East Coast & West Coast & Southern States\\
         \hline
         Density        &0.161     (0.037)$^1$  &0.114     (0.028)$^1$ &0.170     (0.009)$^1$ \\ 
         Unemployment   &0.016     (0.024)      &-0.094    (0.024)$^1$ &-0.007    (0.009) \\
         Income         &0.036     (0.027)      &-0.062    (0.027)$^2$ &-0.043    (0.010)$^1$ \\
         Non-White      &0.096     (0.034)$^1$  &0.226     (0.023)$^1$ &0.160     (0.008)$^1$ \\
         constant       &1.094     (0.019)$^1$  &0.675     (0.019)$^1$ &0.889     (0.007)$^1$ \\
         \hline
         Observations   &3,036              &2,926            &30,668            \\
         \hline
         \hline
         \multicolumn{4}{c}{Phase II}\\
         \hline
         Variables &  East Coast & West Coast & Southern States\\
         \hline
         Density        &0.111     (0.046)$^2$  &0.142     (0.031)$^1$ &0.1006     (0.006)$^1$ \\ 
         Unemployment   &0.039     (0.030)      &-0.083    (0.027)$^1$ &-0.0188    (0.006)$^1$ \\
         Income         &-0.035    (0.034)      &-0.177    (0.031)$^1$ &-0.1103    (0.007)$^1$ \\
         Non-White      &0.038     (0.042)      &0.226     (0.027)$^1$ &0.0067     (0.006) \\
         constant       &1.814     (0.024)$^1$  &1.844     (0.021)$^1$ &2.1862     (0.006)$^1$ \\
         \hline
         Observations   &3,588              &3,458             &36,224             \\
         \hline
         \hline
         no. of counties &138               &133               &1,394              \\
         \hline
    \end{tabular}
    \vspace{0.1cm}
    \caption{\it Results of the causal analyses performed with random effects models. The central value and the robust standard errors (in brackets) are given for the variables that were considered in the analyses.The $p$-values are marked as: $^1 p<0.01$, $^2 p<0.05$.}
    \label{tab:ecotab1}
\end{table}

%#####################
\section*{Comparison between methods}
%#####################
Having used two very different frameworks for this analysis which point towards a very similar interpretation of the data, we would like to highlight the major differences between the frameworks themselves in an attempt to allow the readers to judge the merits of the novel machine learning framework that we propose for causal analyses. 

\subsection*{Non-linearities}
Linear models like the random effect model fail to capture non-linearities in the data and their use is further complicated by correlations present amongst the exogenous variables. This leads to the necessity of considering only those variables that are not highly correlated which leaves open the possibility of ignoring a confounding variable. On the other hand, the framework based on ensembles of BDTs that we use naturally models non-linearities in the data taking into account all correlations and, hence, provide a better regression of the data. This allows for the consideration of a larger set of exogenous variables, which in turn, allows for a better modeling of the endogenous variables.

\subsection*{Interpretability}
The drawback of using machine learning (beyond linear regression or other simple models) is that they are not easily interpretable and often end up as black-boxes. This is, however, not the the case for linear models like the random effects model which allows for a very clear interpretation of the model parameters. The lack of interpretability is, in fact, a major hurdle in the use of machine learning in computational socioeconomics. We address this problem by using Shapley values to interpret the regression model we create with the ensemble of BDTs and do so in a manner in which the casusality in the data is probed, in effect, adding interpretability to a black-box model.

%#####################
\section*{Summary}
%#####################

The work that we present here has two primary components. Firstly, we establish the tenets of a multivariate causal analysis using interpretable machine learning which aims at consistently taking into account non-linearities and correlations and build upon Shapley values, taken from coalition game theory, to extract causal information from the data. Our use of an ensemble of BDTs, instead of other non-linear machine learning models, is motivated by the ease of computation of Shapley values from BDTs using available software packages and should not be taken as a limitation of the causal analysis we propose on a choice of the machine learning models that can possibly be used. The causal Shapley values we calculate are based on the hypotheses about the causal connections between the exogenous variables which are represented by causal chain graphs built on assumptions of the partial causal orderings between the variables. We have tested several hypotheses of these causal orderings to ascertain the robustness of the analysis framework and its dependence on such hypotheses.

Secondly, we use this analysis framework to study the effects of socioeconomic disparities on the spread of COVID-19 in the USA. We use population demographics data gathered from the US Census Bureau and COVID-19 prevalence data from the Johns Hopkins University to fit regression models using the interpretable machine learning framework. The extraction of causal Shapley values from these regression models allow us to infer on the causal connections between the socioeconomic metrics and the prevalence of COVID-19 at the county level. To compare the results of our analysis with a more traditional causal analysis we use a weekly balanced panel data to fit a random effects model with a reduced set of exogenous variables and find reasonable agreement with the results from the interpretable machine learning analysis.

From these analyses we conclude that:
\begin{itemize}
\itemsep0em
    \item The effects of socioeconomic disparities on the spread of COVID-19 was more pronounced at the beginning of the pandemic than at the later stages.
    \item While in the parts of the USA with higher population density, the spread was driven partially by the population density, socioeconomic metrics like the fraction of non-white population in a county also show significant causal connection with the spread of COVID-19. In fact, population density was not causally connected to the spread of the disease in the West coast region.
    \item Of particular note is the causal connection between the fraction of senior citizen in a county and the spread of the disease in the West Coast region and the Southern States. The Shapley values being anti-correlated to the COVID-19 confirmed case rate implies that counties with a younger population saw a larger spread of the disease. As we know that the probability of COVID-19 infection increases drastically with age, our results imply that while the older fraction of the population were being differentially affected to a larger extent by the disease it was being spread more widely by the younger and more mobile fraction of the population.
\end{itemize}

While we have not addressed the question of confounding variables in any detail, it is possible to assume the presence of confounding amongst the variable clustered in a partial ordering. We present this analysis in the Supplementary Information. Assuming the presence of confounding variables does not change the results of our analysis.

With this work, we hope that our proposed methods for causal analysis finds some utility in computational socioeconomics much beyond the application that we have proposed. The conclusions that we draw about the effects of socioeconomic disparities on the spread of COVID-19 in the USA, especially during the onset of the pandemic reinforces what has been observed with clinical data and population level analyses and points to the necessity for restructuring of the crisis response system to nullify the causes of such disparities. We hope our work will lead to more detailed thoughts, insights and actions that will prove to be useful in the near future. 

%%%%%%%%%%%%%%%%%%%%%%%%%%%%%%%%%
% \matmethods{
\section*{Methods}
The data for this work comes from three primary sources:

\begin{itemize}
\itemsep0em
    \item The 2019 \ac{ACS} 5-years supplemental update to the 2011 Census found in the USA Census Bureau database for constructing the socioeconomic metrics.
    \item The population density data that reflects the 2019 estimates of the US Census Bureau.
    \item Data on COVID-19 prevalence and death rate is obtained from the Johns Hopkins University, Center for Systems Science and Engineering database~\cite{Dong:2020ka} sourced from  \href{https://github.com/CSSEGISandData/COVID-19}{github.com/CSSEGISandData/COVID-19}.
\end{itemize}

To study the relevance of socioeconomic condition in the spread of COVID-19, we focus on various metrics that can characterize these at a county level including factors such as per capita income, poverty, the employed fraction of the population, and the unemployment rate. The latter two are not fully correlated since they add up to the fraction of the population that is employable which varies from county to county. We also include factors that relate to the mobility such as the mean commute time in any county along with the fraction of the population that uses a transit system\footnote{We do not consider the relative reduction in mobility due to partial lock-downs or closures of institutions.}. We also include the fraction of senior citizen in a county which can also affect the mobility patterns in the county. For example, a population with a lower median age will, on an average, have higher mobility than a population with a much higher median age. Since studies have already shown that COVID-19 tends to affect the elderly preferentially, the fraction of a population that fall into the senior citizen category (over 65 years of age) is a good measure of quantifying both the mobility and the degree to which age plays a role in determining the spread of COVID-19. To account for professions that put individuals at higher risk of exposure to a larger number of people such as those in service industry, construction, delivery, labor etc., which increases their chance of being infected by COVID-19, we employ a metric that quantifies the fraction of the population that work in these industries per county. Lastly, we include the fraction of individuals in a county that do not have health insurance as a metric to assess if it affects the spread of the disease.

The socioeconomic metrics for each county collected from the US Census Bureau 2019 5-year \ac{ACS} data used in this work are the following:
\begin{itemize}
\itemsep0em
    \item {\bf Population Density (Den)}: The population density data taken from 2019.
    \item {\bf Non-White (NW)}: The fraction of non-white population in any county including Hispanics and Latinos.
    \item {\bf Income (Inc)}: The income per capita as defined by the US Census Bureau.
    \item {\bf Poverty (Pov)}: The fraction of the population deemed as being below the poverty line.
    \item {\bf Unemployment (Uemp)}: The unemployment rate as defined by the US Census Bureau.
    \item {\bf Uninsured (Uins)}: The fraction of the population that does not have health insurance.
    \item {\bf Employed (Emp)}: The fraction of the population that is employed.
    \item {\bf Labour (Lab)}: The fraction of the population working in construction, service, delivery or production.
    \item {\bf Transit (Tran)}: The fraction of the population who take the public transportation system or carpool excluding those who drive or work from home.
    \item {\bf Mean Commute (MC)}: The mean commute distance for a person living in a county in minutes.
    \item {\bf Senior Citizen (SC)}: The fraction of the population that is above 65 years of age.
    \item {\bf Gini index, or Gini coefficient (GI)}: A measure of the distribution of income across a population used as a gauge of economic inequality.
\end{itemize}

\noindent The abbreviations in parentheses are what we use to refer to these variables while discussing partial causal ordering and the results. For the data on COVID-19 prevalence we look at the total number of confirmed cases and deaths up until the 15$^{th}$ of January 2021 and define the following rates:
\begin{itemize}
    \item {\bf Confirmed Case Rate:} The total number of confirmed cases per 100,000 individuals in any county.
    \item {\bf Death Rate:} The total number of deaths per 100,000 individuals in any county.
\end{itemize}
We do not use the fatality rate (number of deaths per confirmed case) as a measure of the disease spread since in~\cite{10.1088/2632-072X/ac0fc7} it was shown the fatality rates are mostly uncorrelated with the socioeconomic metrics pointing to the fact that fatality rates were not disproportionately high in certain socioeconomic strata of the society.

Comorbidities are known to aggravate COVID-19 infections leading to higher chances of a symptomatic infection and hospitalization. We include the data provided in a study conducted by the Centers for Disease Control and Prevention~\cite{10.15585/mmwr.mm6929a1} which provides the distribution of comorbidities in all the counties in the USA to study the possible effects of comorbidities in determining the extent to which COVID-19 spreads in any region. We define the corresponding metric as:
\begin{itemize}
    \item {\bf Comorbidities (Com)}: The total fraction of population with one or more pre-existing chronic conditions.
\end{itemize}

%#####################
\subsection*{Analysis Framework}
%#####################

In this work we aim at extracting the causal connections between the exogenous variables (socioeconomic metrics) and endogenous variables (COVID-19 prevalence and death rate) by using machine learning for modeling underlying multivariate distributions from the data, and then calculating causal Shapley values. The procedure we follow can be delineated as:

\begin{itemize}
\itemsep0em
    \item Use an ensemble of \acp{BDT} which act as weak learners that perform reliably in a statistical ensemble. This allows us to build a non-linear model of COVID-19 case and death rates in terms of the socioeconomic metrics and population density at the county level.
    \item Define the causal flow of the variables by setting up partial causal ordering graphs.
    \item Calculate the causal Shapley values that allow us to infer upon the causal connections between the exogenous variables and the endogenous ones.
\end{itemize}

To clarify the analysis procedures we delve into some details of the machine learning framework that we use and the definition of causal Shapley values. Of particular importance is our discussion of how we set up the partial causal ordering graphs that determine the causal flow of the variables. This is the part where an understanding of the interactions between the variables is necessary and we use several hypothesis to make sure we properly address the subjective nature of designing the partial causal orderings.

% --------------------
\subsection*{Machine learning framework}
\label{sec:ml}
% --------------------
To perform a regression of the data we use an ensemble of \acfp{BDT} taking either disease prevalence or death rate as the endogenous variable while keeping the socioeconomic metrics and population density as exogenous variables. This allows us to not assume a functional form for the model but rely on the data alone. We emphasize here that we are not trying to build a predictive model using machine learning but rather to perform regression in a model-agnostic manner. We use XGBoost~\cite{10.1145/2939672.2939785}, a scalable end-to-end boosting system for decision trees that is particularly suitable for sparse data. For data augmentation we use an ensemble of \acp{BDT} as weak learners trained on random selections of the sample with replacement split $70/30$ into training and testing sets. This also gives us a stable measure of the accuracy with which we can model the data. We use the coefficient of determination, $R^2$, calculated from the test sample set aside for each \ac{BDT}, as our regression accuracy measure and estimate the error of estimating $R^2$ from the ensemble. It is normal for weak learners to differ in their predictions (it is a consequence of bagging within the ensemble). The prediction of an ensemble of weak learners is taken as the average in a regression problem. The accuracy of the ensemble is represented by the mean $R^2$.

% --------------------
\subsection*{Causal Shapley values}
% --------------------
The Shapley value, first introduced in~\cite{Shapley1953}, is a solution concept from cooperative game theory for transferable utility games. The game is characterised by a pair $(C, F)$, where $F=\{1,\ldots,d\}$ is a set representing players in the game, and the \textit{characteristic function} $C:2^F\rightarrow \mathbb{R}$ assigns a non-negative real value $C(S)$ to every coalition $S \subseteq F$, and zero to the empty coalition, i.e. $C(\varnothing) = 0$. The Shapley decomposition is provably (cf.~\cite[Thm. 2]{Young1985}) the only solution concept satisfying the four favorable axioms of \textit{efficiency}, \textit{additivity}, \textit{symmetry} and \textit{null player}, well exposed in~\cite{Huettner2012}. This uniqueness has contributed to the popularity of the Shapley value in the literature of \acf{XAI}. However, different incarnations of Shapley value based explanation methods make different assumptions and use of different approximations, rendering Shapley value based explanations less ``unique''.

Using Shapley values in a machine learning context amounts to interpreting the players $F$ as features in a model, and the characteristic function $C$ as either the model itself or an evaluation measure of the model's performance. Lundberg et al.~\cite{Lundberg2017} identified that several existing explanation methods, including Shapley value based ones, belong to a class of ``additive feature attribution methods'', and unified Shapley values with the solution concept of \ac{LIME}~\cite{LIME}, introducing the \ac{SHAP} model with simplified inputs, i.e.\ inclusion/exclusion of features. As machine models assume an input shape defined during training, evaluating it in the absence of a feature is not possible in general. Hence, the SHAP characteristic function approximates the (counterfactual) model prediction under removal of features via the expected value of the prediction conditional upon the values of the included features. 

In the~\cite{Lundberg2017} iteration of so-called Kernel \ac{SHAP}, this conditional expectation is approximated via the marginal distribution, see eq.\ (11) in~\cite{Lundberg2017}, which amounts to an assumption of feature independence. This is done for computational efficiency, and not inherent to the \ac{SHAP} framework.
The Python \ac{SHAP} package~\cite{SHAPpackagePython}, first released along with~\cite{Lundberg2017}, is maintained. 
In \cite{Loland2019}, Aas et al.\ suggest improving Kernel \ac{SHAP} by obtaining the values of the out of coalition features by conditioning upon the excluded features when calculating the expected values. As the conditional distribution is in general not known, Ref.~\cite{Loland2019} presents four ways of approximating it. One of these is via the empirical distribution of the data, weighing data instances based on a scaled Mahalanobis distance~\cite{Mahalanobis1936}. The authors of Ref.~\cite{Loland2019} refer to this as conditional Kernel \ac{SHAP}, which they implement in the R-package shapr~\cite{SHAPpackageR}.

In Ref.~\cite{janzing2020}, Janzing et al.\ argue that unconditional, rather than conditional, expectations provide the right notion of excluding features, which contradicts the theoretical justification of~\cite{Lundberg2017}, used in the \ac{SHAP} software packages. Drawing from Pearl's seminal work on causality~\cite{Pearl1995CausalDF}\footnote{The interested reader is referred to~\cite{docalculus} for an introduction to the \textit{do}-calculus}, Janzing et al.\ stress that the \textit{interventional} distribution is represented by the marginal distribution. Leaning heavily on the work of Datta et al.~\cite{datta2016}, Janzing et al.\ thus conclude that the marginal distribution -- not the conditional distribution -- should be used for excluded features in the \ac{SHAP} calculation. 

Shortly after, Heskes et al.~\cite{Heskes2020} show that this marginal, respectively interventional, \ac{SHAP} calculation only represents direct effects, meaning that `root causes' with strong indirect effects are ignored in the feature attribution. They introduce \textit{causal} Shapley values by explicitly including the causal relationships between the data, in the \ac{SHAP} value calculation. They provide an extension to the shapr package, where the user specifies the causal ordering as well as possible confounding, available at~\cite{causal-shapr-package}\footnote{This patch has not yet been included in the official shapr package.}.

In our work, we will use the form of causal Shapley values defined by Heskes et al.~\cite{Heskes2020} through the definition of partial causal ordering graphs. On one hand, this makes the analysis of the causal connection somewhat subjective and based on our decision of the causal connections between the variables (and the presence of confounding variables). On the other hand, this enforcement of the causal ordering can be seen as a ``prior'' with which we inform the computation of the causal Shapley values the information that might not be directly gleaned from the data. We will, in particular, use two quantities to examine the data. The first one is the mean of the absolute Shapley values, $\overline{|S_v|}$, that serves as a measure of global importance of a variable~\cite{Lundberg2020}. Higher the value of  $\overline{|S_v|}$ of a variable, the greater effect the variable has in determining the outcome. We shall also define $S_I$ as the integer index corresponding to the position in a hierarchy that a certain variable holds when ordered by the corresponding $\overline{|S_v|}$ with $S_I\in [1,n]$ for a $n$-variable problem.

%############################
\subsection*{Data Availability}
%############################

All the data used in this work is sources from public databases that have been cited in the main text. All codes that have been used to process and analyze the data can be found at \href{https://github.com/talismanbrandi/Causal-Inference-IML-C19}{https://github.com/talismanbrandi/Causal-Inference-IML-C19}.

% \clearpage
\bibliography{bibliography}
% \bibliography*{}

\section*{Acknowledgements}
A.P. is funded in part by Volkswagen Foundation within the initiative ``Corona Crisis and Beyond -- Perspectives for Science, Scholarship and Society'', grant number 99091.
I.\ S.\ is grateful for support received through the EXAIGON project from industrial partners and the Research Council of Norway (grant no.\ 304843).
This research was supported in part through the Maxwell computational resources operated at DESY, Hamburg, Germany.

\section*{Author contributions statement}
\label{sec:ac}
All authors contributed in their own capacities in performing the analysis and writing the draft. The contribution of V.S. deserves a special mention. He contributed significantly to the conceptual design and implementation of the causal analysis, the building of the causal orderings and data analysis.

% \section*{Competing interests}
% The authors declare that they have no competing interests.

\includepdf[pages=-]{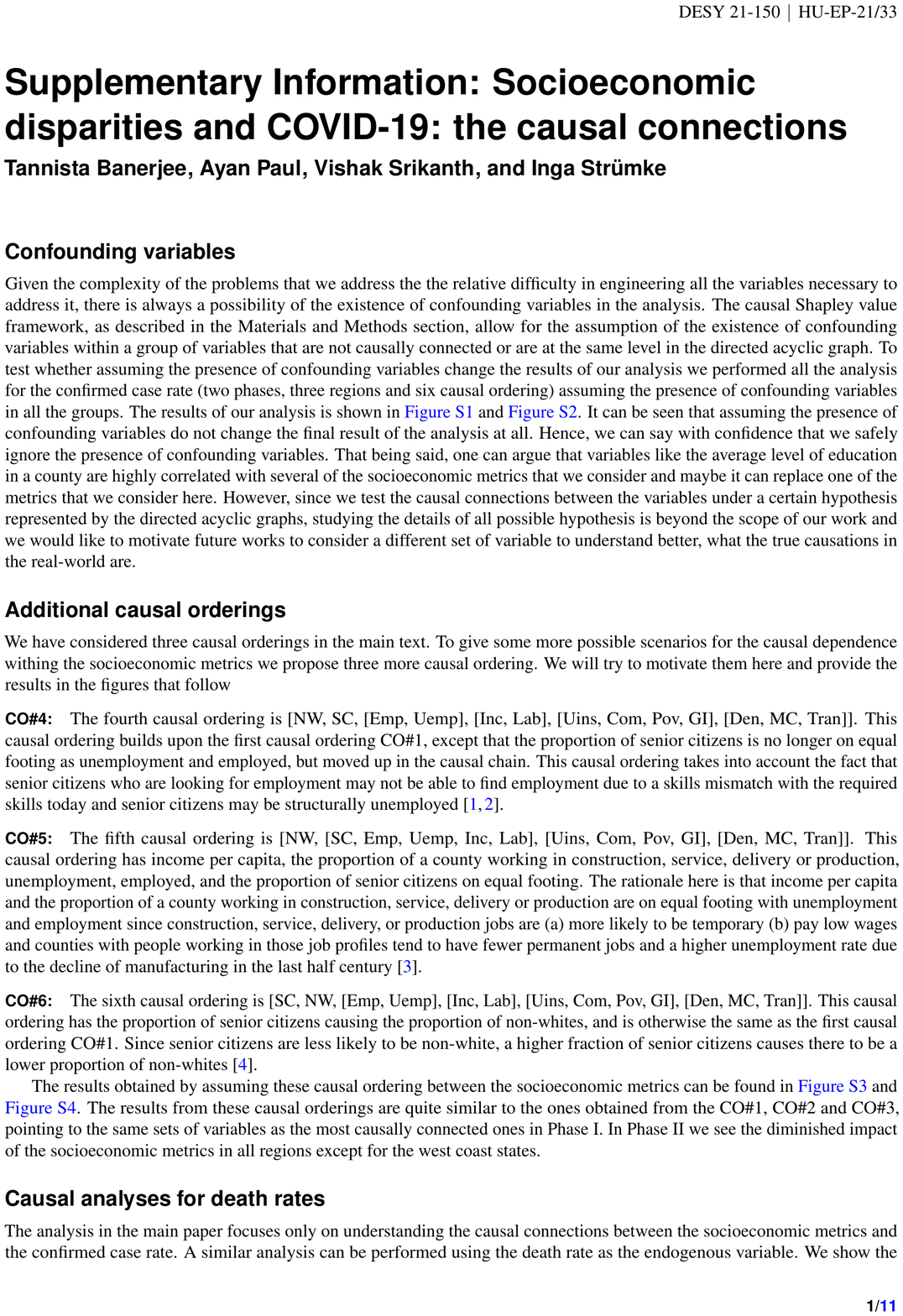}

\end{document}